\documentclass[conference]{IEEEtran}
\IEEEoverridecommandlockouts
\usepackage{amsmath, amssymb, amsfonts}
\usepackage{algorithmic}
\usepackage{graphicx}
\usepackage{textcomp}
\usepackage{xcolor}
\usepackage[
backend=biber,
style=authoryear,  
sorting=ynt,
natbib=true        
]{biblatex}

\usepackage{caption}
\usepackage{subcaption}  
\usepackage[export]{adjustbox}
\usepackage{multirow} 
\usepackage{tikz}
\usepackage{hyperref}
\usepackage{pgfplots}
\pgfplotsset{compat=1.17} 

\def\BibTeX{{\rm B\kern-.05em{\sc i\kern-.025em b}\kern-.08em
    T\kern-.1667em\lower.7ex\hbox{E}\kern-.125emX}}
\begin{document}

\title{Variable Stiffness for Robust Locomotion through Reinforcement Learning*\\
{\footnotesize \textsuperscript{*}accepted to the 16th IFAC joint symposia of mechatronics and robotics}
}

\author{\IEEEauthorblockN{Dario Spoljaric}
\IEEEauthorblockA{\textit{Autonomous Systems} \\
\textit{TU Wien}\\
Vienna, Austria \\
dario.spoljaric@tuwien.ac.at}
\and
\IEEEauthorblockN{Yan Yashuai}
\IEEEauthorblockA{\textit{Autonomous Systems} \\
\textit{TU Wien}\\
Vienna, Austria \\
yashuai.yan@tuwien.ac.at}
\and
\IEEEauthorblockN{Dongheui Lee}
\IEEEauthorblockA{\textit{Autonomous Systems, TU Wien} \\
\textit{Institute of Robotics and Mechatronics, DLR}\\
Vienna, Austria \& Wessling, Germany \\
dongheui.lee@tuwien.ac.at}
}

\maketitle

\begin{abstract}
Reinforcement-learned locomotion enables legged robots to perform highly dynamic motions but often accompanies time-consuming manual tuning of joint stiffness.
This paper introduces a novel control paradigm that integrates variable stiffness into the action space alongside joint positions, enabling grouped stiffness control such as per-joint stiffness (PJS), per-leg stiffness (PLS) and hybrid joint-leg stiffness (HJLS).
We show that variable stiffness policies, with grouping in per-leg stiffness (PLS), outperform position-based control in velocity tracking and push recovery. In contrast, HJLS excels in energy efficiency. 
Despite the fact that our policy is trained on flat floor only, our method showcases robust walking behaviour on diverse outdoor terrains, indicating robust sim-to-real transfer.
Our approach simplifies design by eliminating per-joint stiffness tuning while keeping competitive results with various metrics.
\end{abstract}


\section{Introduction}
Animal and human-like locomotion has a significant advantage over wheeled mobile robots. They can traverse unstructured, challenging terrains. Therefore, various approaches are developed to solve quadrupedal and bipedal locomotion. Conventionally, this involved model-based controllers \citep{kim2019highly, di2018dynamic} with a complex pipeline that managed gait schedule, state estimation, whole body impulse control and actuator control.
Recently, this discipline has made significant progress through model-free reinforcement learning (RL) methods. These approaches enable the design of controllers capable of following high-level commands (walking, running, jumping, etc.) and directly actuating the joint motors, bypassing the need for path planning and other parts within the control pipeline.

Usually, these controllers \citep{torque_quadro, genloco} follow a position- or torque-based paradigm.
The high-level controller (RL agent) learns a position policy in the position-based paradigm. Given the state, the RL agent predicts desired joint positions at low frequencies, which are transferred into torques by a high-frequency PD controller. This control paradigm requires manual engineering of motor stiffness and damping for different tasks.  
In contrast, humans and animals can adjust their stiffness and damping to handle different tasks. For example, we stiffen our foot joints when landing with a foot but relax in the swing phase.
The torque-based control circumvents this by directly learning a torque policy, which shows higher compliance \citep{torque_biped}. Generally, torque-based policies are more challenging to train because they require learning complex dynamics.
On the other hand, position control has a good initial pose, easing the learning progress in the beginning. 
Torque-based RL agents are usually executed at higher speeds, necessitating more powerful hardware or smaller networks. 
This leaves the researcher with a choice, either additional tuning of joint stiffnesses or a hard-to-train policy with limits in scope. 
\begin{figure}
    \centering
    \includegraphics[width=0.95\linewidth]{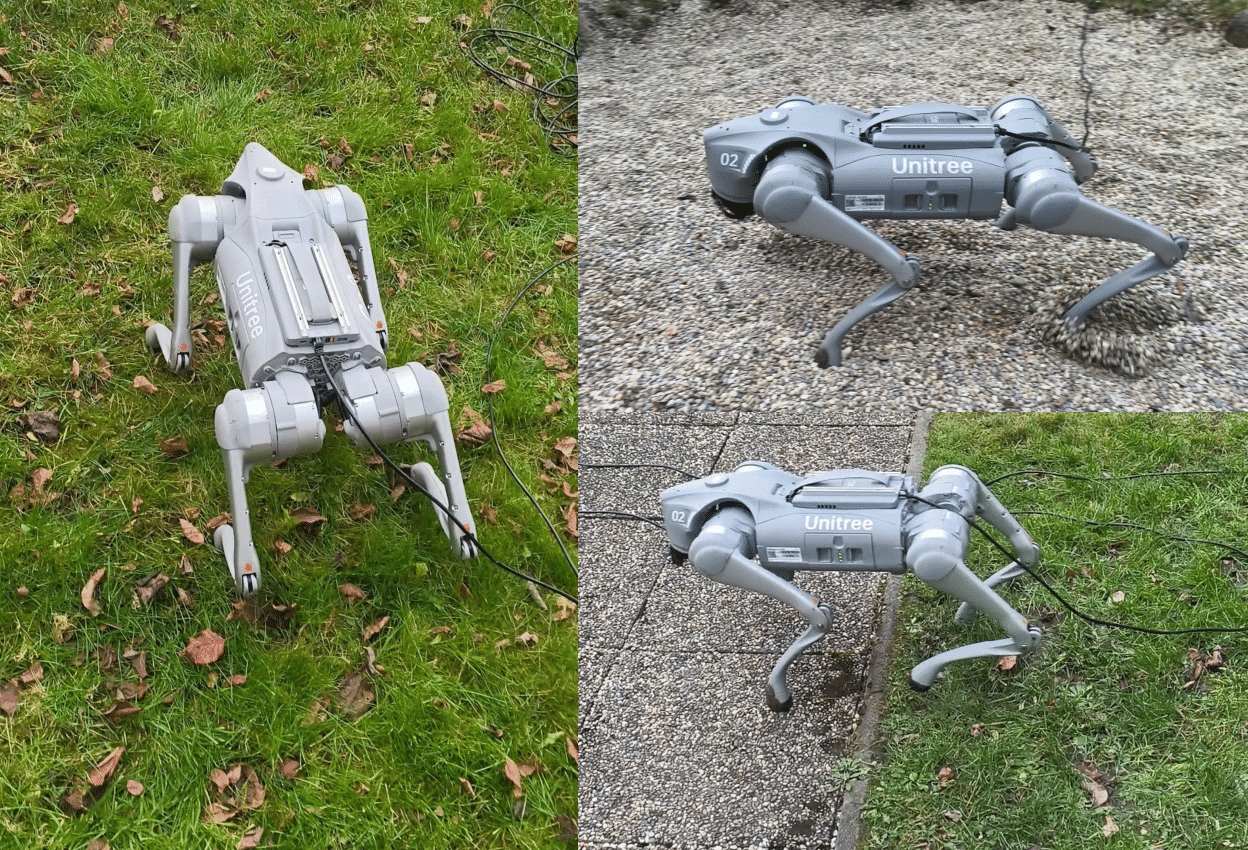}
    \caption{A quadrupedal robot traverses across different terrains with variable-stiffness RL policy.}
    \label{fig:vic_control}
\end{figure}

Another approach \citep{yu2024compact, fu2024novel} is variable stiffness control, which is widely used in robotics, particularly in manipulator applications, to improve safety while still being able to track accurately. Preliminary studies from \cite{yu2024compact} have shown that applying this technique to locomotion could enhance energy efficiency. This also seems reasonable as higher stiffness is required during contact with the ground, but less is needed during the swing phase.
In quadruped locomotion, research from \cite{zhao2022variable} within model-based frameworks has shown that variable stiffness can reduce contact forces.  However, successfully achieving locomotion with variable stiffness remains a significant challenge. 

This paper studies reinforcement-learned controllers that learn joint stiffness alongside target positions. 
Consequently, the robot can automatically adjust the motor stiffness according to the task requirements. Our approach shows superior velocity tracking and push recovery performance while maintaining good energy efficiency and robust sim-to-real transfer performance. Results of the walking policy are shown in Fig. \ref{fig:vic_control} and a video is available online\footnotemark. 
\footnotetext{\url{https://drive.google.com/file/d/1SmwQSM5026Ri41Ue6J_IgP0NqYIFxCOt/view?usp=sharing}}
\begin{figure*}
    \centering
    \includegraphics[width=0.7\textwidth]{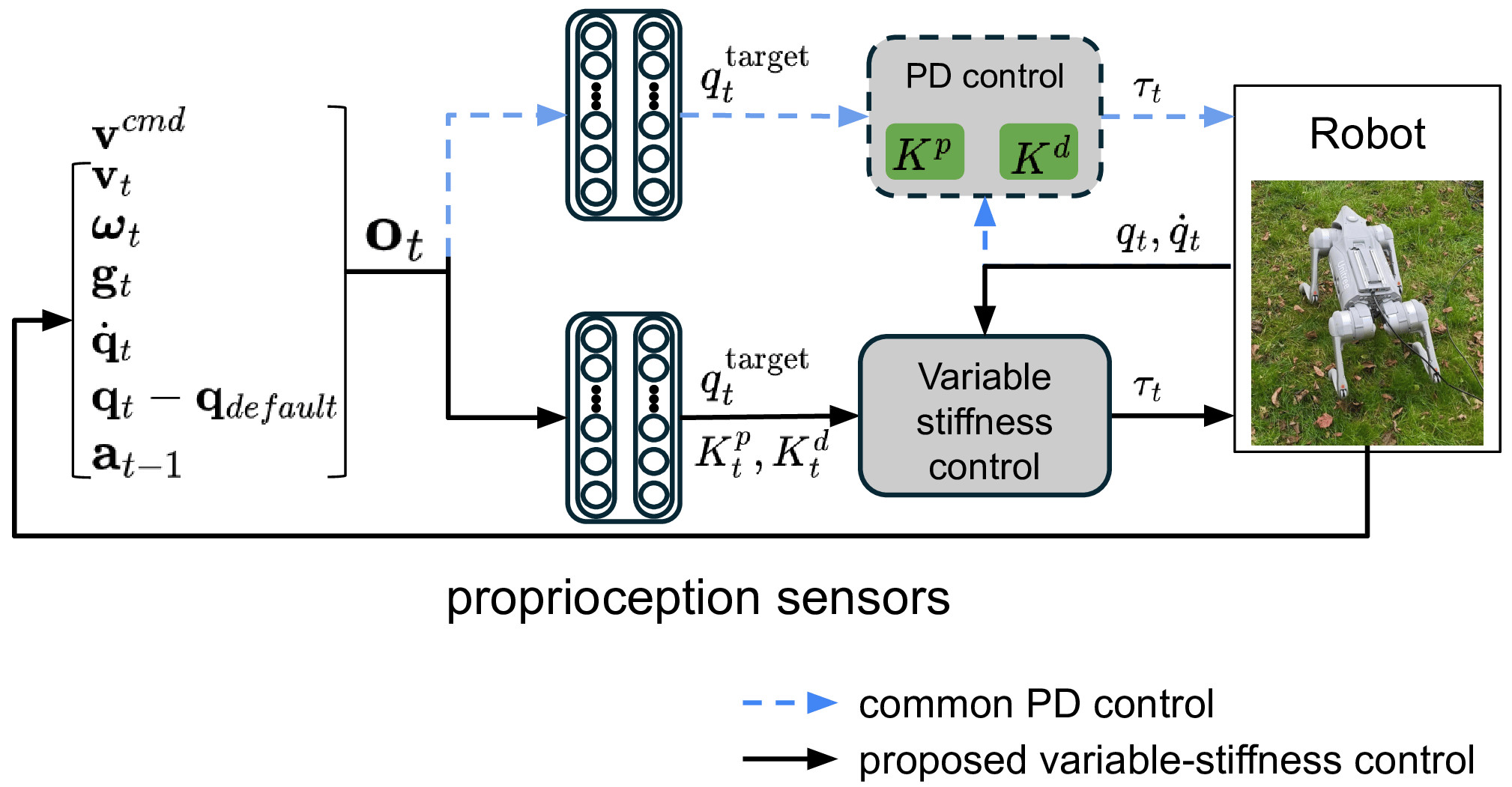}
    \caption{Architecture of the position-based control (blue dashed line) compared to our variable stiffness control.}
    \label{fig:control_concept}
\end{figure*}

\section{Related work}
RL has emerged as a promising approach for solving locomotion tasks, offering two primary paradigms: position-based and torque-based control.
\cite{genloco} and \cite{dreamwaq} predict target joint positions, which are actuated using proportional-derivative (PD) controllers. These methods are widely favoured for their ease of training and robustness in high-level command tracking. 
Within this control, the torque applied to the motors is calculated using Eq. (\ref{eq:action_torque})
\begin{equation}
\tau_t = K_p (q_{t}^{target}(\theta)- q_t) + K_d ( \dot{q}_{des} - \dot{q_t} ) \label{eq:action_torque}
\end{equation}
However, they suffer from limited compliance behaviour due to fixed stiffness and damping values (\(K_p\) and \(K_d\)), requiring extensive PD gain tuning that is often task- and robot-specific \citep{genloco}. 

In this control, the PD controller acts as a low-level tracking module, where the proportional gain (P-gain) is representative of the stiffness and the derivative gain (d-gain) for the damping. \cite{9560837} investigated the impact of the p-gains, suggesting that large proportional gain leads to instabilities in training. In contrast, the low proportional gain has significant tracking errors and behaves like a torque controller. Other research studied the impact on the derivative gain. \cite{smith2023demonstrating} showed small derivative gains result in learning instabilities, and large gains prevented tracking the target velocity. 

To circumvent this PD controller, \cite{torque_quadro} and \cite{torque_biped} studied torque control as an alternative and applied it to quadruped and biped locomotion. In this control, the actions are directly applied to the motors.
Although this control showed higher achievable rewards in the long term, it must be executed at higher speeds to perform similarly to position control. It is more difficult to train initially. The higher control speeds limit the design freedom of torque-based controllers. 

\cite{zhao2022variable} showed for model-based control, that adapting stiffness according to the contact force led to sufficient walking for a quadruped on uneven terrains. \cite{vic_sensible_tasks} studied the impact of including joint stiffness alongside joint positions. 
The torque was then calculated by Eq. (\ref{eq:variable_stiffness}).  
\begin{equation}
    \tau_t = K_t^p(\theta) (q_{t}^{target}(\theta) - q_t) - K_t^d(\theta) \dot{q_t}  \label{eq:variable_stiffness}
\end{equation} 
Overall, their work did not test their approach to locomotion but showed the superiority of this concept against torque and position control for two scenarios: a single-legged hopper and a manipulator. The single-legged hopper achieved more considerable jumping heights, and the manipulator maintained a continuous contact force during reference motions with the adjustment of the action space. 

Inspired by these findings, we explore the integration of joint stiffness alongside joint positions in the action space for reinforcement learning-based quadruped locomotion.

\section{Methods}
We aim to train a policy $\pi_{\theta}$ with parameters $\theta$ that can follow high-level velocity command $\mathbf{v}^{cmd}=[\mathbf{v}_{xy}^{cmd}, \omega_{yaw}^{cmd}]^T$. This includes the lateral velocity $v_{xy} ^{cmd}$ as well as angular rotation speed $\omega_{yaw} ^{cmd}$. The prediction of joint targets along joint stiffnesses should accomplish this. An overview of our approach is shown in Fig. \ref{fig:control_concept}.
\subsection{Training}
We train our controllers using Proximal Policy Optimisation (PPO) \citep{ppo} with 4096 environments in parallel for 2000 epochs. 
To improve learning efficiency, we apply early termination if the robot's orientation exceeds 90 degrees from its horizontal position, if joint or torque limits are exceeded or if the robot falls onto its hips, trunk, or LIDAR. The episode lasts 20 seconds, and we sample commands every 5 seconds.

\textbf{Simulation environment:}
We use Mujoco-MJX (introduced by \cite{mujoco}) as the simulation environment and apply Domain randomisation to achieve a successful sim-to-real transfer.
Table \ref{tab:rando} shows the randomised parameters.
\begin{table}
\centering
\begin{tabular}{cccc}
\hline
\textbf{Term} & \textbf{Distribution} & \textbf{Units} & \textbf{Operator} \\
\hline
\multicolumn{4}{c}{\textbf{Environmental properties}} \\
\hline
Payload Mass (trunk) & $\mathcal{U}(-1.0, 3.0)$ & kg & additive \\

Hip Masses & $\mathcal{U}(-0.5, 0.5)$ & kg & additive \\
Ground friction & $\mathcal{U}(0.3, 1.25)$ & - & multiplicative \\
Gravity offset & $\mathcal{U}(-1.0, 1.0)$ & m/s² & additive \\
\hline
\multicolumn{4}{c}{\textbf{Noise in the observation space}} \\
\hline
Joint positions & $\mathcal{U}(-0.01, 0.01)$ & rad & additive \\

Joint velocities & $\mathcal{U}(-1.5, 1.5)$ & rad/s & additive \\

Local velocity & $\mathcal{U}(-0.1, 0.1)$ & m/s & additive \\

Local ang. Velocity & $\mathcal{U}(-0.2, 0.2)$ & rad/s & additive \\

Projected gravity & $\mathcal{U}(-0.05, 0.05)$ & rad/s² & additive \\

System delay & $\mathcal{U}(0.0, 15.0)$ & ms & additive \\

Stiffness & $\mathcal{U}(0.8, 1.3)$ & N/m & multiplicative \\

Damping & $\mathcal{U}(0.5, 1.5)$ & kg/s & multiplicative \\

Motor strength & $\mathcal{U}(0.9, 1.1)$ & - & multiplicative \\

\hline
\end{tabular}
\caption{Domain randomisation in simulation}
\label{tab:rando}
\end{table}
Additionally, we expose the policies to external pushes applied from random xy directions.
The force magnitudes of the pushes are randomised between 50 - 150 N, and the impulse is 8 - 15 Ns. The push is applied every 6 seconds, so the policies have to learn to react to such disturbances.

\textbf{Observation:}  
Observations passed to the actor must also be observable during deployment. This limits the freedom of the observations that are passed. The observation vector $\mathbf{o}_t$, as in Eq. (\ref{eq:observation}), is composed of the body angular $\boldsymbol{\omega}_{t}$ and linear velocity $\mathbf{v}_t$ , projected gravitational vector $\mathbf{g}_t$, joint angle difference from the default position $\mathbf{q}_t- \mathbf{q}_{default}$, the last action $\mathbf{a}_{t-1}$ and the velocity command  $\mathbf{v}^{cmd}$. 
\begin{equation}
    \mathbf{o}_t = \left[\mathbf{v}^{cmd}, \mathbf{v}_t ,\boldsymbol{\omega_{t}}, \mathbf{g}_t, \dot{\mathbf{q}}_t, \mathbf{q}_t - \mathbf{q}_{\text{default}}, \mathbf{a}_{t-1} \right]^T
    \label{eq:observation}
\end{equation}

We utilize privileged information to learn the critic network, which is dropped in the inference phase. The privileged vector $\mathbf{s}_t$ consists of the scaling factor for proportional and derivative gains and motor strength $\sigma_t$. In addition, $\mathbf{s}_t$ contains ground friction, adapted masses, disturbance force, and regular observation, as stated in Eq. (\ref{eq:priv_observation}).
\begin{equation}
 \mathbf{s}_t = \left[k_{p,t}, k_{d,t}, \sigma_t, \mu,m_t, \mathbf{F}_{\text{kick}}, \mathbf{o}_t \right]^T    \label{eq:priv_observation}
\end{equation}

\textbf{Reward:}
The reward functions, similar to \cite{dreamwaq, rudin2022learning, vic_sensible_tasks}, consist of task rewards to follow the command given and auxiliary rewards (penalties) to stabilize the learning process. An overview is provided in Table \ref{tab:rewards}. 
\begin{table}
\centering
\resizebox{\columnwidth}{!}{
\begin{tabular}{|l|l|l|}
\hline
Reward & Equation ($r_i$) & Weight ($w_i$) \\
\hline
Lin. velocity tracking & $\exp\left(-4\left(\mathbf{v}_{xy}^{\text{cmd}} - \mathbf{v}_{xy}\right)^2\right)$ & 1.5 \\
Ang. velocity tracking & $\exp\left(-4\left( \omega_{\text{yaw}}^{\text{cmd}} - \omega_{\text{yaw}}\right)^2\right)$ & 0.8 \\
Linear velocity (z) & $v_z^2$ & -2.0 \\
Angular velocity (xy) & $\boldsymbol{\omega}_{xy}^2$ & -0.05 \\
Orientation & $\mathbf{g}_{xy}^2$ & -5.0 \\
Feet air time & $\sum_{f=0}^4(t_{air,f}-0.1), |\mathbf{v}_{cmd}| > 0.1 $ & 0.2 \\
Joint accelerations & $\ddot{\boldsymbol{q}}^2$ & $-2.5 \times 10^{-7}$ \\
Joint power & $ \boldsymbol{|\tau|} \cdot |\dot{\boldsymbol{q}}|$ & $-2 \times 10^{-5}$ \\
Power distribution & $\text{var}(|\boldsymbol{\tau}|\cdot |\dot{\boldsymbol{q}}|)$ & $-10^{-5}$ \\
Foot slip & $\sum_{f=0}^{4} ||\mathbf{v}_{f, xy}||^2, \text{if } z  < 0.01$ & -0.1 \\
Action rate & $(\mathbf{a}_t - \mathbf{a}_{t-1})^2$ & -0.01 \\
Foot clearance & $\sum_{f=1}^4\left( \mathbf{p}_{f,z}^{des} - \mathbf{p}_{f,z}\right)^2 |v_{f,xy}|$ & -0.1 \\
Center of mass & $ \left(\mathbf{p}_{com,xy} - \mathbf{p}_{xy}^{des}\right)^2,\ \mathbf{p}_{xy}^{des} = \frac{\sum_{f=1}^4 \mathbf{p}_{f,xy}}{4}$ & -1.0 \\
Joint tracking & $ (\mathbf{q}_t^{target} - \mathbf{q}_{t+1})^2$ & -0.1 \\
Base height & $(h^{des}-h)^2$ & -0.6 \\
Hip & $\exp \left(-4 * \sum_{k=1}^{4}(q_{hip,k}-q_{hip,k}^{default} )^2 \right)$ & 0.05 \\
Collisions & $n_{collisions}$ & -10.0 \\
Termination & $n_{termination}$ & -10.0 \\
\hline 
\end{tabular}
}
\caption{Reward functions.}
\label{tab:rewards}
\end{table}

Given the reward components, the total reward is calculated using Eq. (\ref{eq:reward_calculation}). 
\begin{equation}
r_t = \sum r_i \cdot w_i \cdot dt
\label{eq:reward_calculation}
\end{equation}
In addition to existing rewards from the literature, we introduced a "Center of Mass" reward to encourage a stable walking gait. This reward calculates the desired centre-of-mass position $p_{xy}^{des}$, as the mean xy-components of the feet positions and penalises the squared error from this target, ensuring the centre of mass remains within the support polygon.

\textbf{Action Space:}
The position-based controller uses 12 actions to specify target positions for each joint. Our controllers extend actions to adjust the stiffness of the PD controller, ranging from 20 to 60. To study the stiffness adjustments on quadrupedal robots with 12 DoFs, we propose different grouping strategies:
\begin{itemize} 
\item \textbf{Individual Joint Stiffness (IJS)}: Predicts stiffness for each joint individually, extending the action space to 24 dimensions.
\item \textbf{Per Joint Stiffness (PJS)}: Groups joints into hip, thigh, and knee categories, predicting one stiffness per group for a 15-dimensional action space.
\item \textbf{Per Leg Stiffness (PLS)}: Predicts one stiffness value per leg, adding four actions for a 16-dimensional space.
\item \textbf{Hybrid Joint-Leg Stiffness (HJLS)}: Combines PJS and PLS by representing stiffness as the outer product of a leg stiffness vector ($\mathbf{k^l} \in \mathbb{R}^4$) and a joint group stiffness vector ($\mathbf{k^j} \in \mathbb{R}^3$), resulting in 19 dimensions. 
\end{itemize}
The damping of the PD controller is set to a fixed relationship to the p-gain, according to Eq. \ref{eq:damping_ratio}.
\begin{equation}
K_t^d =  0.2 \sqrt{K_t^p}
\label{eq:damping_ratio}
\end{equation}
This relationship is inspired by the ratio between PD gains as in the works of \cite{genloco} and \cite{rudin2022learning}.

\section{Evaluation}

In this section, we present the experimental results in both Mujoco-MJX and real-world hardware experiments to answer the following questions:
\begin{itemize}
\item Can variable stiffness policies improve velocity command tracking performance?
\item Do variable stiffness policies show greater robustness against disturbances?
\item Are variable stiffness policies more energy efficient?
\end{itemize}

\subsection{Baseline}
The baseline is formed by position control policies with low stiffness 20, referred to as P20, and high stiffness 50, referred to as P50. These baselines, trained under identical conditions, follow prior choices from \cite{dreamwaq, margolis2024rapid} for lower stiffness and \cite{genloco} for higher stiffness. While \cite{genloco} used even higher gains, we observed instabilities in training and selected 50 as the upper stiffness. These baselines highlight trade-offs, with each stiffness excelling in specific tasks.

\subsection{Performance on walking and running}
\label{sec:walking_n_running}

\begin{figure}
\centering
\begin{tikzpicture}
\begin{axis}[
    ybar,
    enlargelimits=0.15,
    legend style={at={(0.5,-0.25)},
      anchor=north,legend columns=6
      },
      legend image code/.code={
        \draw[#1] (0cm,-0.1cm) rectangle (0.3cm,0.1cm); 
    },
    ylabel={Mean tracking error [m/s]},
    symbolic x coords={0.5 m/s, 0.8 m/s, 1.0 m/s},
    xtick=data,
    width=0.48\textwidth, 
    height=0.24\textwidth, 
    bar width=6pt, 
]
\addplot coordinates {(0.5 m/s,0.0411) (0.8 m/s,0.04704) (1.0 m/s,0.0505)};
\addlegendentry{IJS}
\addplot coordinates {(0.5 m/s,0.0158) (0.8 m/s,0.01966) (1.0 m/s,0.0294)};
\addlegendentry{PJS}
\addplot coordinates {(0.5 m/s,0.0132) (0.8 m/s,0.01448) (1.0 m/s,0.0191)};
\addlegendentry{PLS}
\addplot coordinates {(0.5 m/s,0.0232) (0.8 m/s,0.038254) (1.0 m/s,0.0582)};
\addlegendentry{HJLS}
\addplot coordinates {(0.5 m/s,0.049) (0.8 m/s,0.0833) (1.0 m/s,0.0834)};
\addlegendentry{P20}
\addplot coordinates {(0.5 m/s,0.0215) (0.8 m/s,0.02509) (1.0 m/s,0.0253)};
\addlegendentry{P50}
\end{axis}
\end{tikzpicture}

\caption{Absolute tracking error for trained policies.}
\label{fig:combined_speed_tracking}
\end{figure}
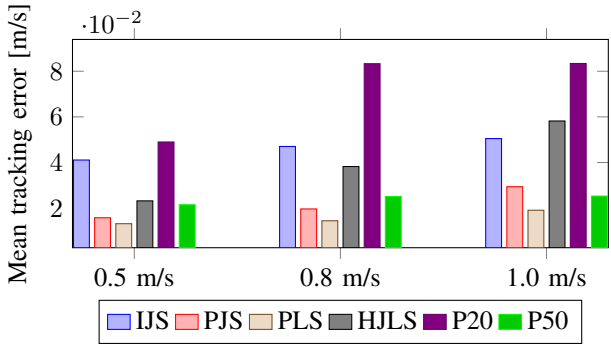

We evaluate walking and running performance by measuring tracking errors between the commanded and achieved velocities, a common method used in the works of \cite{Lee_2020}. Controllers are tested on eight discrete headings (0°, 45°, ..., 315°) with target speeds of 0.5 m/s, 0.8 m/s, and 1.0 m/s to prevent bias for speed class or direction. 
Each heading direction is held for eight seconds, and domain randomisation is turned off to focus on tracking accuracy.
Fig. \ref{fig:combined_speed_tracking} shows the absolute velocity tracking error averaged over the heading directions at different speeds. The comparison indicates that P20 has a significant tracking error, whereas P50 maintains lower error. 
Predicting individual joints in a policy (IJS) demonstrates the highest tracking error. However, our grouped stiffness policies, PJS and PLS, show lower or comparable tracking errors than P50.
PLS manages to outperform all other controllers in every speed class.

\subsection{Push recovery}
Robustness is evaluated by exposing the locomotion policy to external disturbances. Specifically, force pushes to the robot's trunk with domain randomisation. The evaluation is performed in the same simulation environment as the training.
Push recovery is measured under the following conditions: 
\begin{itemize} 
\item Walking speed: 0.3 m/s 
\item Force push magnitude: 50 - 300N 
\item Push duration: 0.1 sec 
\end{itemize} 
Pushes are applied randomly in the xy-plane. The robot must walk straight for 5 seconds, with a random push applied between 2.5 and 3.5 seconds. A fall results in failure, while recovery and walking for 5 seconds are considered a success.
The randomisation of the push event is done to prevent bias due to specific postures.

The success rate for pushes within the magnitude constraints is reported in Table \ref{tab:force_push_comparison}. Within the push force of $150N$ (training range), our controller PJS performs best. Above $150 N$ IJS shows the highest success rate, closely followed by PLS and HJLS. 
Since PLS also outperformed the baselines in velocity tracking and shows the second-highest success rate ($<300N$), we use this policy for further comparison.

\begin{table}
\centering
\resizebox{\columnwidth}{!}{
\begin{tabular}{|c|c c c c c |}
\hline
\multirow{2}{*}{Control paradigm} & \multicolumn{5}{c|}{Success Rate (\%)} \\ 
\cline{2-6}
 & $<100N$ & $<150N$ & $<200N$ & $<250N$ & $<300N$ \\
\hline
IJS & \textbf{100.00} & 98.73 & 96.06 & \textbf{93.53} & \textbf{90.44} \\
PJS & \textbf{100.00} & \textbf{100.00} & 96.63 & 89.34 & 81.64 \\
PLS & \textbf{100.00} & 99.66 & \textbf{97.66} & 92.34 & 85.65 \\
HJLS & 99.83 & 99.32 & 97.03 & 91.58 & 83.93 \\
P20 & 99.48 & 99.07 & 94.98 & 89.93 & 83.08 \\
P50 & \textbf{100.00} & 99.75 & 97.20 & 89.93 & 81.77 \\
\hline
\end{tabular}
}
\caption{Push recovery success rates within the specified push force magnitudes.}
\label{tab:force_push_comparison}
\end{table}

\begin{figure}[!t]
\centering
    \begin{minipage}[b]{0.49\linewidth}
        \centering
        \includegraphics[width=\linewidth]{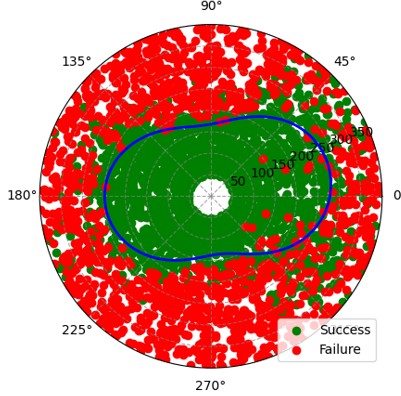}
        \subcaption{}
        \label{fig:scatter_plot}
    \end{minipage}%
    \hfill
    \begin{minipage}[b]{0.49\linewidth}
        \centering
        \includegraphics[width=\linewidth]{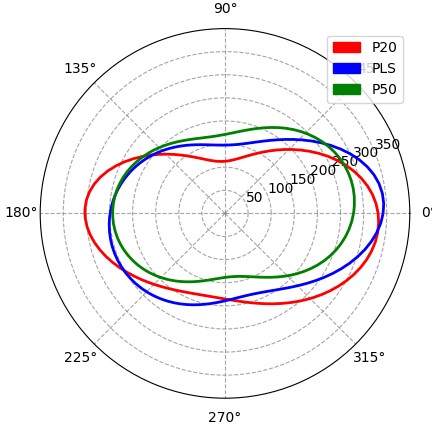}
        \subcaption{}
        \label{fig:push_comparison_pos_vic}
    \end{minipage}%
    
     \caption{\textbf{Comparison of maximal recoverable force push.}
     The polar scatter plot (a) shows the outcome of the experiment for specific forces (red=failure, green=success). The SVM, with classification confidence 90\%, is used to draw a success boundary in the polar plot (a), which is then used to compare our methodology against the baselines in plot (b).}
    \label{fig:comparison_force_push}
\end{figure}

For comparison, we draw a polar scatter plot illustrating the magnitude and angle of the force applied to the robot trunk. We utilise a Support vector machine(SVM) with a radial bias function kernel to classify a maximum recovery boundary. This maximum recovery boundary is used to compare the methods. 
Figure \ref{fig:comparison_force_push} shows the results of this experiment. 
\begin{figure}
    \centering
    \includegraphics[width=0.8\linewidth]{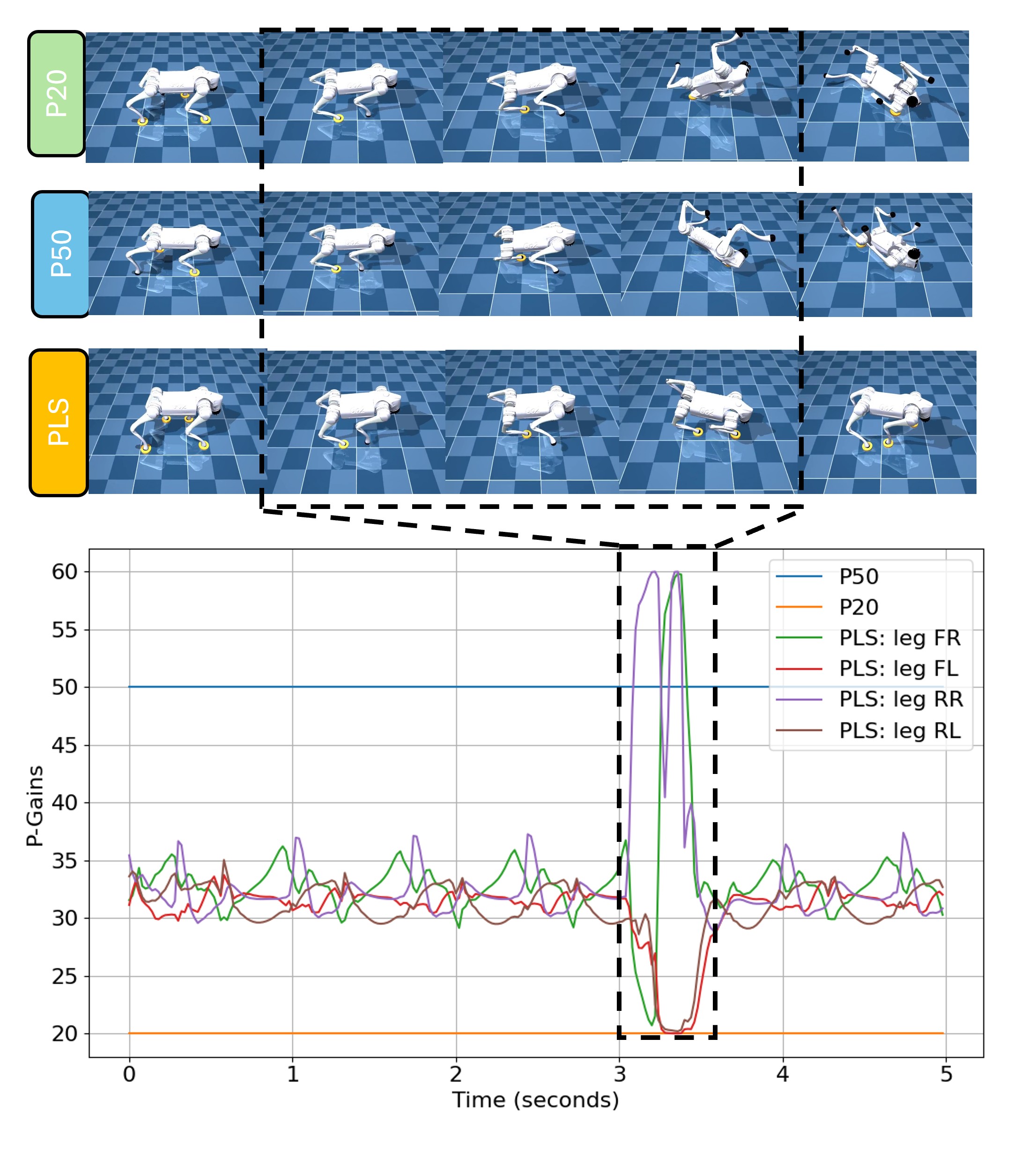}
    \caption{\textbf{Push experiment}: Stiffness plotted over time. Push conditions: $t_{push}=3 s$, $t_{duration}=0.1 s$, $F=190 N$, $\theta = 225^\circ$ accounting for the point in time, duration, magnitude and direction at which the push is applied. The force is applied from the frontal left direction. (FR/FL: front left/right, RR/RL: Rear right/left)}
    \label{fig:push_test_stiff}
\end{figure}

All policies show higher resilience in and against the walking direction (0, 180°). Pushes applied from the side are not well compensated. P50 shows a smaller region than P20. Our policy PLS demonstrates a convex and similar shape to the P20.

Figure \ref{fig:push_test_stiff} further demonstrates how the baseline policies react to a force of 190N applied from the frontal left direction. The graph shows the stiffness values plotted over time for the respective legs.
The baseline policies fall, whereas the PLS manages to recover by adjusting the stiffnesses of the legs. Notably, the legs opposite to the push stiffen way above the maximum stiffness of the high stiffness policy, and the legs towards the push relax. These results demonstrate the superior performance of variable stiffness in this setting.

\subsection{Energy efficiency}

Similar to the work of \cite{Lee_2020}, we evaluate energy efficiency with the cost of transport (CoT) for different speed classes. They define the CoT as Eq. (\ref{eq:cot}), where $M_{total}$ accounts for the total mass of the robot, $\tau$ denotes the measured torques, $\dot{q}$ the joint velocities, $g$ the gravitational acceleration and $v$ for the measured velocity. 

\begin{equation}
CoT = \frac{E}{M_{total}gd} = \frac{P}{M_{total}gv} = \frac{\tau \dot{q}}{M_{total}gv}
\label{eq:cot}
\end{equation}

\begin{figure}
\centering
\begin{tikzpicture}
\begin{axis}[
    ybar,
    enlargelimits=0.15,
    legend style={at={(0.5,-0.25)},
      anchor=north,legend columns=6
      },
      legend image code/.code={
        \draw[#1] (0cm,-0.1cm) rectangle (0.3cm,0.1cm); 
    },
    ylabel={Cost of Transport},
    symbolic x coords={0.5 m/s, 0.8 m/s, 1.0 m/s},
    xtick=data,
    width=0.48\textwidth, 
    height=0.24\textwidth, 
    bar width=6pt, 
]
\addplot coordinates {(0.5 m/s, 0.6342) (0.8 m/s, 0.6971) (1.0 m/s, 0.7334)};
\addlegendentry{Model PJS}

\addplot coordinates {(0.5 m/s, 0.654) (0.8 m/s, 0.6885) (1.0 m/s, 0.7168)};
\addlegendentry{PLS}

\addplot coordinates {(0.5 m/s, 0.7972) (0.8 m/s, 0.7952) (1.0 m/s, 0.807)};
\addlegendentry{IJS}

\addplot coordinates {(0.5 m/s, 0.5944) (0.8 m/s, 0.6627) (1.0 m/s, 0.709)};
\addlegendentry{HJLS}

\addplot coordinates {(0.5 m/s, 0.5988) (0.8 m/s, 0.6433) (1.0 m/s, 0.6914)};
\addlegendentry{P20}

\addplot coordinates {(0.5 m/s, 0.7116) (0.8 m/s, 0.7508) (1.0 m/s, 0.7722)};
\addlegendentry{P50}
\end{axis}
\end{tikzpicture}

\caption{Cost of Transport for trained policies.}
\label{fig:cot}
\end{figure}
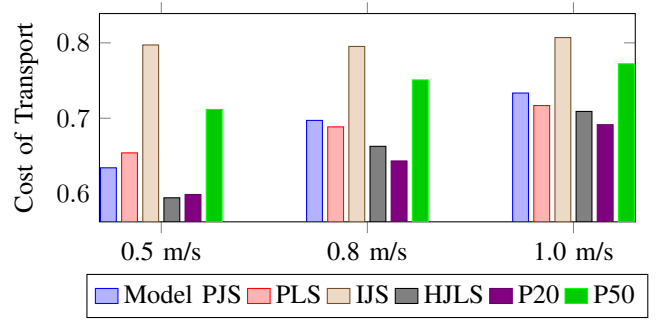

We measure this metric applied to the same experiment as described in section \ref{sec:walking_n_running}. The results are shown in Fig. 
\ref{fig:cot}
and reveal mixed results. The lower the CoT, the more energy efficient. 
Our policies PJS, PLS and HJLS demonstrate lower CoT than P50 but higher than P20. This is also expected as higher stiffness leads to higher torques and, therefore, higher energy consumption. Our policies can adjust their stiffness between 20 and 60; thus, we expect the CoT to be somewhere in between. Lowering the lower border of stiffness compromised gait quality in our experiments.
HJLS demonstrates the lowest CoT among our policies and, therefore, remains competitive with P20. These results validate the effectiveness of our proposed methodology in grouping the stiffness.

\subsection{Sim-to-real transfer} 
\begin{figure}
\centering
\includegraphics[width=0.49\textwidth]{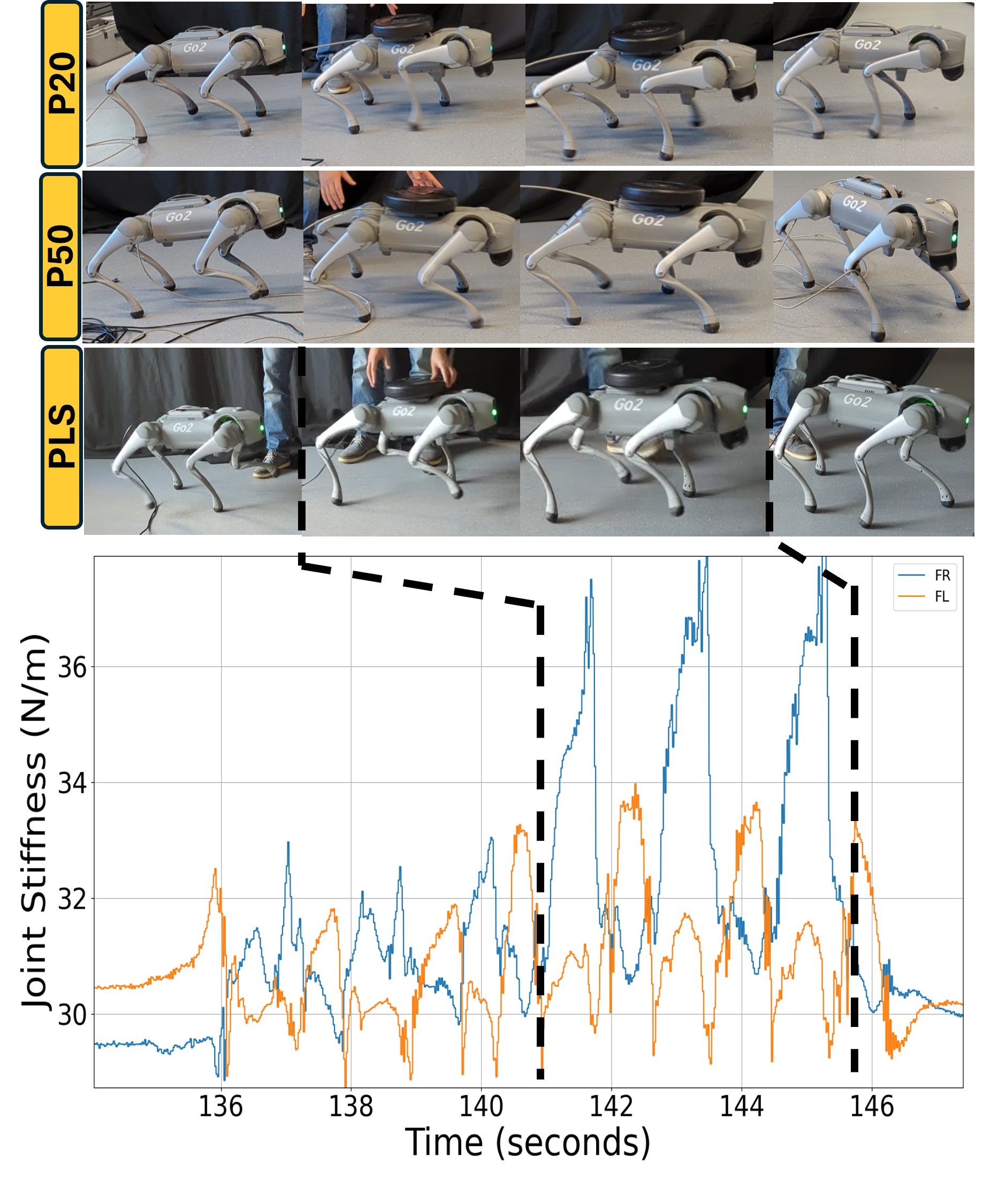}
\caption{\textbf{Payload experiment:} Adding a 5kg mass during walking to the baselines and PLS policy.}
\label{fig:payload_experiment}
\end{figure}

\begin{figure*}[!h]
    \centering
    \includegraphics[width=0.9\textwidth]{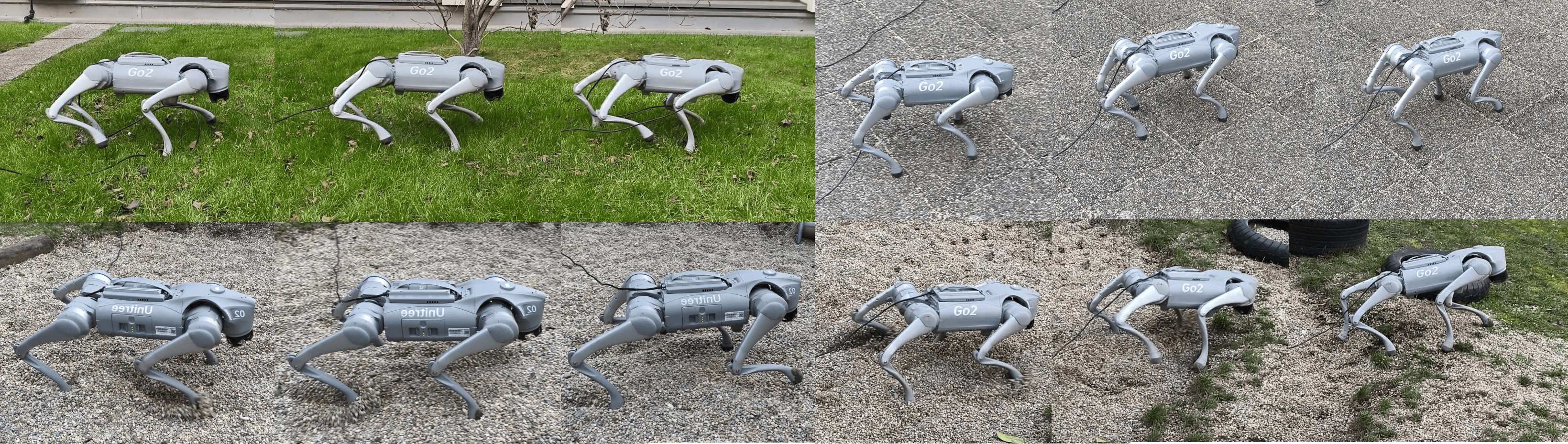}
    \caption{Despite being trained solely on a flat floor in simulation, we showcase the robot's ability with our variable-stiffness RL policy to walk robustly on diverse outdoor terrains, such as grass, stones, and sand.}
    \label{fig:outdoor_experiments}
\end{figure*}

During hardware deployment, we evaluate the robustness of the learned locomotion by adding a payload or walking over diverse terrains. In training, the policy encounters a randomised payload of up to 3kg. For evaluation, we add a 5 kg payload on the robot during walking and observe increased stiffness while maintaining a proper gait, as seen in Fig. \ref{fig:payload_experiment}. When the payload is removed, the stiffness decreases accordingly.

We also evaluate our variable stiffness policy in the outdoors to traverse various terrains. Although it is trained solely on a flat floor, the learned policy demonstrates robust walking across diverse surfaces, including mud, grass, sidewalks, and sand (see Fig. \ref{fig:outdoor_experiments}).

\section{Future work}
Our work demonstrates the benefits of variable stiffness. In future works this method could be applied to learn even more different tasks like crouching, hopping stair traversal and imitating motions. As this approach is able to adjust the stiffness this method might also learn these tasks for multiple robot types and combine it into one policy. 

\section{Conclusion}
In this paper, we studied an alternative approach to learning locomotion on a quadruped robot, which uses joint positions alongside stiffnesses as the action space in a reinforcement learning paradigm.
Simulation and real-world experiments are conducted to investigate performance on walking and running, as well as push recovery, energy efficiency and sim-to-real transfer. Our policy, which predicts stiffness per leg, outperformed baselines in the robustness test and velocity tracking. On the other hand, individual joint stiffness prediction struggled, underscoring the efficiency of our found groupings. 
Hardware tests show stiffness adaptation when encountered with payload and robust walking over diverse terrains. 
Our research highlights the potential of reinforcement learned variable stiffness locomotion as it combines the advantages of low and high stiffness.


\printbibliography
\end{document}